# Robotic Backchanneling in Online Conversation Facilitation: A Cross-Generational Study

Sota Kobuki[1], Katie Seaborn[1], Seiki Tokunaga[2], Kosuke Fukumori[2], Shun Hidaka[1], Kazuhiro Tamura[2], Koji Inoue[3], Tatsuya Kawahara[3], and Mihoko Otake-Mastuura[2]

[1]Tokyo Institute of Technology, [2]RIKEN, [3]Kyoto University

# Robotic Backchanneling in Online Conversation Facilitation: A Cross-Generational Study

Sota Kobuki[1], Katie Seaborn[1], Seiki Tokunaga[2], Kosuke Fukumori[2], Shun Hidaka[1], Kazuhiro Tamura[2], Koji Inoue[3], Tatsuya Kawahara[3], Mihoko Otake-Mastuura[2]

([1]Tokyo Institute of Technology, [2]RIKEN, [3]Kyoto University)

*Abstract*—Japan faces many challenges related to its aging society, including increasing rates of cognitive decline in the population and a shortage of caregivers. Efforts have begun to explore solutions using artificial intelligence (AI), especially socially embodied intelligent agents and robots that can communicate with people. Yet, there has been little research on the compatibility of these agents with older adults in various everyday situations. To this end, we conducted a user study to evaluate a robot that functions as a facilitator for a group conversation protocol designed to prevent cognitive decline. We modified the robot to use backchannelling, a natural human way of speaking, to increase receptiveness of the robot and enjoyment of the group conversation experience. We conducted a cross-generational study with young adults and older adults. Qualitative analyses indicated that younger adults perceived the backchannelling version of the robot as kinder, more trustworthy, and more acceptable than the non-backchannelling robot. Finally, we found that the robot's backchannelling elicited nonverbal backchanneling in older participants.

## I. Introduction

Japan is experiencing societal aging, which is characterized by a rapidly increasing population of older adults alongside falling birth-rates. In parallel, the global prevalence of dementia is also increasing rapidly. However, there is not enough support for the number of patients. In response, new strategies are being explored in Japan and abroad. One is artificial intelligence (AI). Notably, there are growing expectations for AI-based intelligent agents to interact with many people and bring various benefits. One of the most promising ways to deal with the problem of dementia is to use human-centered AI to help slow down the decline of cognitive functions that decline with dementia [1]–[3]. Such AI systems may entertain people and provide companionship as well as have anti-dementia effects. But they must be easy to use and appropriate for older adults.

We explored "Bono Bot," an intelligent robotic agent designed for older adults that facilitates small group conversation using an elicitation method called "Coimagination" [4]. In feedback from our previous face-to-face studies [5], participants expressed a desire for empathic responses like backchannelling from the agent. So, we evaluated the user experience (UX) and usability of a new version of Bono Bot that uses *backchannelling*: the vocal interjections people make during conversation to indicate that they are engaged. Specifically, we compared "with backchannelling" versus "without backchannelling" versions of the robot. While planning this research, the COVID-19 began, tremendously affecting social interactions for people of all ages by limiting face-to-face interactions and obscuring face-to-face non-verbal communication due to mask usage. We were forced to conduct the study on Zoom, which allowed us to evaluate the UX and usability of the *online* context.

Older adults, like everyone, may benefit from new technologies. Even so, they have often been unable to integrate such agents and other novel digital offerings into their lives with ease, an issue known as the "digital divide" [6]. We thus created a system that would be simple and applicable to all ages. We evaluated the system in two stages. First, we involved young people to evaluate the system's baseline usability and UX, especially ease of use, ease of speaking, and enjoyment. Next, we involved older adults, conducting a comparative user study to extend the findings cross-generationally and explore a backchannelling-using robot. We asked the following three questions:

**RQ1: What usability and UX does the online version of the group conversation method facilitated by a robot using backchannelling provide?**

**RQ2: How does the robot's use of backchannelling relate to participants' ease of speaking?**

**RQ3: How does the robot's use of backchannelling relate to participants' enjoyment and sense of fun?**

Our main contributions are: (i) knowledge of how conversational facilitation robots that use backchannelling can have a positive influence on group conversation, even online; cross-generational design findings on robotic backchanneling; and a proof-of-concept prototype made up of the robot, elicitation system, and the online environment. This work is a first step towards enabling age-inclusive positive and natural social interactions with online conversational robots.

## II. Procedure for Paper Submission

### A. Robots and Group Conversation

Many have explored group conversation with robots, with work on robots being modeled to take part in human-to-human group conversations going back twenty years [7]. A variety of studies about group conversation robots have been done since. Mutlu et al. conducted research on a robot that uses gaze cues to establish its own and its partner's roles in a conversation, called "footing" [8]. Fujie et al. conducted research in which a robot participated in a quiz with older adults at a day-care center, finding that the robot succeeded in entertaining these older adults [9]. In general, research on how robots can participate in group conversation with people has been conducted for many years [10]. Other applied experiments are also being conducted, such as research on the line of sight of the robot or the android in group conversation experiments with robots, and further development is desired [11], [12]. These findings show that the relationship between robots and group conversation have a lot of promise and potential. Nevertheless, online contexts are understudied, and specific features of robotic social embodiment [5] and communication, such as backchannelling, require more attention.

### B. Robots and Backchanneling

Backchannelling is a phenomenon that is frequently observed in conversations and is considered to be a very important element in facilitating con-versation [13]. It plays a role in indicating that the listener is listening, understanding, and sympathizing with the speaker [14], as well as in creating the rhythm of the conversation. Backchannelling is often expressed as "umm-umm" or "ahh" sounds in Japanese.

Many conversational systems using backchannelling have been studied [15], [16]. Here we used an implementation of the system created by Kawahara and colleagues at Kyoto University [17]. In general, backchannelling has been found to lead to more user friendly and positive experiences [18], [19]. For instance, one study described a simple talking robot with backchannel feedback that is designed based on artificial subtle expressions (ASE). These ASE-based expressions of backchannel feedback provided a positive experience [20]. Based on this, we predict that backchannelling will improve users' experiences even in online situations. This is represented in our first two hypotheses:

*H1: Usability scores (SUS) related to robotic facilitation will be improved by the inclusion of backchannelling.*

*H2: The backchannelling version of the robot will contribute to a more positive atmosphere compared to the non-backchannelling version.*

Other research has shown that people wanted to talk to a backchannelling-using robot again after using it [17]. Based on this, we predicted that backchannelling can increase people's desire to use the robot again.

*H3: The backchannelling version of the robot will increase desire to use the robot again compared to the non-backchannelling version.*

Other research has found that when robots used back-channeling, team performance improved, and the robots were seen as more engaged [19]. We thus predicted that the backchannelling-using robot would be perceived as more interested and be listening keenly, even in online situations, we made the following hypothesis:

*H4: The backchannelling version of the robot will be perceived as a better listener compared to the non-backchannelling version.*

### III. SYSTEM DESIGN

We used two versions of Bono Bot, an intelligent agent for facilitating group conversation: (i) "with backchannelling" and (ii) "without backchannelling." Both versions of the agent used the same conversation-parsing AI for the "Coimagination" activity, which is a group conversation method for older adults. We describe in detail below.

### A. Conversational Format: Coimagination Method

Coimagination is a conversation support method defined by the following two rules: (1) Each participant brings a topic along with materials such as photographs, illustrations, music, and real objects according to a predetermined theme. (2) The order and duration of the presentations are determined. There is a set time for the presentation of the topic and a set time for questions and answers. The speaker should concentrate on speaking and the listener should concentrate on listening while the speaker's picture or material is projected. The roles of listeners and speakers are switched each time, thus providing equal opportunity for all participants to speak, listen, ask questions, and respond. The Coimagination Method aims to activate three cognitive functions that decline at an early stage of mild cognitive impairment, episodic memory, division of attention, and planning function. We chose the Coimagination method because the method has been established as an intervention for older adults and it can be used for group conversation. Moreover, Coimagination, unlike other protocols, had a speech-interpreting AI platform "Fonopane." We could use it as the baseline intelligent agent.

### B. Robot: Bono Bot

Bono Bot (Fig. 1) is an AI-based intelligent agent with a semi-humanoid robotic form that facilitates small group conversation using the Coimagination Method. It has been developed over several years. It can react to the voices of participants and turn towards them. It can also move its arms to make gestures. Since this study was conducted online, it could not face the direction of the participants, so we set the robot to face the direction of the camera showing it.

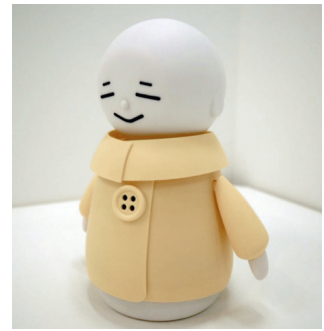

Figure 1. Bono Bot, the group conversation facilitation robot.

### C. Facilitating Software: Fonopane

"Fonopane" is used in Coimagination sessions to present the photos taken by participants. It can be projected or shown by monitor; we showed it to participants via Zoom screen sharing. The main functions are showing photos and time.

### D. Online Environment: Zoom

Zoom is a cloud-based video communication app that supports collaborative functions. Due to the influence of COVID-19, it has been used more and more in lectures at universities and company meetings but also by the average person in their private lives, to maintain contact with others and join group events safely. We asked participants to have a group conversation through this tool (Fig. 2).

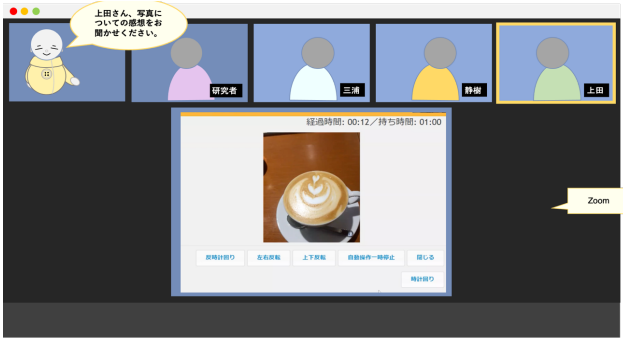

Figure 2. Illustration of the study setup over Zoom.

## IV. STUDY 1: YOUNGER PEOPLE

### A. Methods

*1) Research Design*

A between-subjects study was conducted to achieve the aims of this research. A human factors engineering approach using mixed methods data collection and focusing on usability and UX was taken. The between-subjects factor was the presence or absence of backchannelling in the robot's speech. The groups, to which participants were randomly assigned, were "with backchannelling" and "without backchannelling."

*2) Measures and Measurements*

Usability and UX was captured by questionnaire. Usability consisted of three factors: effectiveness, efficiency, and satisfaction (ISO 9241-11). For these, we used the Japanese version of the SUS (System Usability Scale) [21]. All measures were rated on a 5-point scale from 0 to 4, organized so that higher numbers indicate more positive results. For UX, we used open-ended questions like "How did you feel about participating in the online session facilitated by the robot?"

*3) Participants*

We recruited 18 Japanese students (3 women, 15 men, none of another gender) aged 20+. We only recruited participants who could use Zoom. Also, because the robot used Japanese backchannelling, recruitment was limited to people who could understand conversational Japanese (roughly JLPT N2 level). None had used Bono before. All had a high school education or above. This study was approved by the IRB.

*4) Procedure*

A week before the session, participants were given two themes that they had to take a photo about: "a favorite food" and "a hobby." Note that participants had to take multiple photos on different themes. Before the session, participants provided the photos to the tech staff, who uploaded them. On the day of the study, they entered the Zoom meeting room. They were greeted and received an explanation about the study. Then they filled out consent forms for the study and the session began with an introduction by the research assistant about the Coimagination Method and Bono, and then facilitation was passed to Bono. They had two 15–20-minute sessions. And finally, they filled out questionnaires about the studies. Participants were given compensation for this study.

*5) Data Analysis*

Quantitative and qualitative data analyses were used. We used descriptive and inferential statistics for evaluating individuals and groups across the quantitative measures which were collected by questionnaire. We used a thematic analysis approach to the open-ended responses. For this, we coded the data and created an initial set of themes based on how these codes could be categorized together. We also counted how frequently the themes appeared.

### B. Results

Participants were asked about the pace of conversation and ease of speaking, but no significant results were obtained. Thus, in the quantitative results, H1~H4 were rejected.

For the qualitative results, four themes were identified and defined to capture patterns of experience, attitudes, and behavior: "Appreciability" (with: found in 78.9% of the data, without: found in 22.2% of the data), "Enjoyability" (with: found in 44.4% of the data, without: found in 11.1% of the data), "Trustworthiness" (with: found in 66.7% of the data, without: found in 22.2% of the data), and "Positivity" (with: found in 66.7% of the data, without: found in 33.3% of the data). Definitions and examples are given in Study 1.

TABLE I. STUDY 1 THEMES WITH DEFINITIONS AND EXAMPLES.

| Theme | Definition | Example |
|---|---|---|
| Appreciability | The "without aizuchi" version received a larger amount of feedback with negative content. | "I was expecting it to give us a speech or something, but it did not. I felt that a text-only moderator would have been better." |
| Enjoyability | Aizuchi enhanced the enjoyment of the participant's experience. | "It felt fresh and fun." |
| Trustworthiness | Aizuchi enhanced the reliability of the participants. | "It looked so heartwarming, and I felt I could trust it." |
| Positivity | More positive experiences for participants in the "with aizuchi" sessions. | "I think it would be quicker to have a human moderator for this amount of speech." |

### C. Discussion

The findings on Bono Bot's backchannelling during online group conversation were mixed. However, in analyzing the results from multiple perspectives, we found a slightly positive effect of the backchannelling version for younger people.

Most hypotheses were rejected, with one indicating a negative impression of backchannelling. Yet, the qualitative analyses indicate positive UX. The Appreciability and Positivity themes suggest that backchannelling reduced negative feedback and a desire for improvement. The Trustworthiness theme indicated that backchannelling can elicit feelings of trust. In answer to RQ1, backchannelling can have a positive impact even while further improvements are needed to make it less frightening and more humanlike.

Based on the qualitative and quantitative analyses, we were not able to obtain any clear results regarding RQ2. Even so, the Enjoyability theme suggests that the answer to RQ3 is that the backchannelling approach was truly enjoyable. Also, the Appreciability theme indicates that without backchannelling, younger people's impressions were more negative and longer lasting, and led to more suggestions for improvement. This means that previous findings on backchannelling leading to a more user-friendly and positive experience, which various studies have shown [18], [19], also applies to the online context. Nevertheless, regarding "ease of speaking," our results were not as good as those of studies that used the same backchannelling system that was used in this study. It seems that face-to-face contexts may still be ideal on this metric.

*1) Limitations*

The most important limitation of this study was the communication environment. This is because in this study, the amount of participant speech was analyzed in real time, and it affected accuracy of the backchannelling. In fact, some participants pollinated the problem of time lag. Another limitation was the number of participants was small. This study was the first opportunity for online implementation and aimed to prepare for future studies with older adult participants. If we had a larger number of subjects, smaller significant differences might be easier to spot.

## V. STUDY 2: OLDER ADULTS

### A. Methods

We carried out a study with older adults that was similar in design to Study 1. Here we describe the differences in detail.

*1) Research Design*

A within-subjects study was conducted to compare robot versions with a focus on usability and UX. The within-subjects factor was the presence or absence of backchannelling. Subjects were randomly assigned to "with " and "without" backchannelling groups. The order of the conditions was counterbalanced to account for order effects.

*2) Measures and Measurement*

Since the participants in this study were older adults, a paper-based questionnaire was used and the number of questions was reduced to be less burdensome. The questionnaire consisted of the SUS and custom items created by us on the ability of the robot to listen to the speaker. These were: Q2: "Enjoyment of the experience", Q3: "Whether the robot was listening to", Q4: "Whether the robot was fitting into the conversation", and Q5: "Whether they would like to be listened to again." These were answered on a 5-point Likert scale. We also asked two open-ended questions: "What were your impressions of this experience?" and "Do you have any recommendations to improve this experience?" We only collected the qualitative open-ended responses at the end, so that participants could reflect on both conditions. Finally, participants were asked about their experience with online video calls on a 5-point scale (5. almost every day, 4. frequently, 3. occasionally, 2. not often, 1. never).

*3) Participants*

We recruited 24 Japanese older adults (15 women, 9 men, none of another gender) aged 65+ with at least a high school education. As in Study 1, recruitment was limited to those who could understand conversational Japanese. Participants could not use hearing aids, as earphones were required during the study. Only those without dementia were recruited. Some had experience participating in multiple Coimagination sessions facilitated by Bono-Bot. But those were all in-person, so this study was the first time online. Participants infrequently used video conferencing tools like Zoom (M = 2.4, SD = 1.1).

*4) Procedure*

Because of the effects of environmental factors found in Study 1 and the fact that the participants were older adults, we conducted the study by sending each older adult a PC via snail mail. A week before the session, participants were also provided with two topics about which they had to take photos: "Favorite Foods" and "Hobbies." Prior to the session, participants provided their photos to the technical staff, who uploaded them to the Fonopane system. On the day of the study, there was a training session and a main session. In the training session, participants entered the Zoom room and were given an explanation of the study, as well as time to test and ask questions about their technical setup. They then filled out the consent form and received instructions regarding the use of Zoom in the session. Facilitation was then handed over to Bono Bot for the main session, which consisted of two 15- to 20-minute Coimagination events. After each Coimagination event, participants filled out the quantitative questionnaire on the experience. After the second event, participants also filled out the qualitative items about the whole experience. Participants were then compensated and debriefed.

*5) Analysis*

We used descriptive and inferential statistics for evaluating individuals and groups across the quantitative measures. Missing values were compensated by the mean.

### B. Results

*1) Usability (H1 and Exploratory Analyses)*

A paired t-test indicated no statistically significant difference in overall SUS ratings between the backchanneling (M=68.5) and non-backchanneling (M=68.7) versions, $p = .46$. Thus, we must reject H1: Usability of robotic facilitation were not improved by the inclusion of backchannelling.

We also conducted exploratory analyses. A paired t-test indicated a statistically significant difference in terms of need for support (from the SUS) between the non-backchanneling version (M=2.1, SD=1.5) and the backchanneling version (M=1.7, SD=1.2), $t(24) = 1.93$, $p = .03$. Also, a paired t-test indicated a statistically significant difference in terms of ease of learning (from the SUS) between the non-backchanneling version (M=3.0, SD=1.2) and the backchanneling version (M=2.8, SD=1.2), $t(24) = 2.01$, $p = .03$. This suggests that backchanneling lessened a perceived need for support.

*2) Enjoyment (H2)*

A paired t-test indicated no difference in enjoyment ratings between the backchanneling (M=4.6) and non-backchanneling (M=4.6) conditions, $p = .42$. Thus, we must reject H2.

*3) A Better Listener (H3 and H4)*

A paired t-test indicated no difference in the item for whether the robot was listening to between the backchanneling (M=3.4) and non-backchanneling (M=3.2) conditions, $p = .15$. A paired t-test indicated no difference in the item for whether the robot was fitting into the conversation between the backchanneling (M=3.2) and non-backchanneling (M=3.0) conditions, $p = .29$. A paired t-test indicated no difference in the item for whether participants want to be listened to between the backchanneling (M=3.4) and non-backchanneling (M=3.3) conditions, $p = .26$. Thus, we must reject H3: The backchannelling version did not increase a desire to use the robot again. We must also reject H4: The backchannelling version of the robot was not perceived as a better listener compared to the non-backchannelling version.

*4) Prompting Backchannelling (Exploratory Analyses)*

We classified three types of backchannelling: (a) Non-verbal, with movement but no sound; (b) Minimal response, with a short sound but no words, e.g., "un"; and (c) Listener feedback, with one or more words that have meaning, e.g., "oh yeah." We compared the frequency of each type (Table II).

TABLE II. STUDY 2 DESCRIPTIVE STATISTICS FOR BACKCHANNELING, AVERAGED ACROSS PARTICIPANTS.

| Type of Backchanneling | With | Without |
|---|---|---|
| Non-verbal* | M=4.6, SD=4.2 | M=3.5, SD=3.5 |
| Minimal response | M=0.5, SD=1.0 | M=0.9, SD=1.2 |
| Listener feedback | M=0.2, SD=0.4 | M=0.2, SD=0.4 |

* $p < .05$.

A paired t-test indicated a statistically significant difference in frequency of the Non-verbal type between the backchanneling and non-backchanneling conditions, $t(20) = 2.08$, $p = .03$. Still, a paired t-test indicated no difference in frequency for Minimal response and Listener feedback between conditions, $p = .11$ and $p = .5$. This suggests that the robot prompted participants to respond nonverbally unawares.

*C. Discussion*

No hypothesis was statistically supported. Also, the presence of backchannelling itself did not necessarily have a significant effect. Still, the learning and support needs SUS results indicate that the backchannelling version was easier to use. However, there was otherwise no difference based on the presence or absence of backchannelling. This suggests that a robot that uses backchannelling in an online context with older adults aided learnability somehow. While we did not focus on learning, this could have implications for onboarding older adults and for robots that act as learning coordinators, which can be explored in future work, cross-generationally and within and outside of online contexts.

We found that the nonverbal backchanneling performed by participants seemed to be influenced by the presence or absence of the robot's backchanneling. Participants naturally synchronized their own nonverbal backchanneling behaviours with the backchanneling of the robot.

Although not directly related to the research question, we should also mention the difficulty of conducting online studies with older adults. In this study, the tasks to be performed only during the backchannelling version were as follows: (i) with the Zoom meeting open, tap the file icon on the home screen of the PC to start; (ii) press the "Join" button; and (iii) return to the Zoom meeting. Even with fewer operations, many participants found it difficult. This was especially so for those who were not familiar with PCs. There were many barriers indicating the "digital device" is still a challenge that we must consider, especially when conducting experiments in an online environment. We will need to devise ways to overcome them, such as with more training sessions on basic technology use or investing the resources in creating a "one-click" application that launches and sets up all required tech for the study.

*1) Limitations*

Most enjoyed the experience of talking with others, making it difficult to distinguish between "with" and "without" backchannelling versions. The relatively small number of participants is a major limitation. As with Study 1, because the study was conducted online, the accuracy of the backchannelling was affected by the environment of each participant. PCs were sent and set up uniformly to provide some consistency in context, but they were affected by external factors, such as sound outside the building. Also, due to Zoom's noise-canceling feature, it was impossible for us to confirm the presence of any sounds around the participants.

## VI. DISCUSSION SYNTHESIS

The younger adults in Study 1 appreciated the online interactions with the backchanneling version of Bono Bot, finding them enjoyable, trustworthy, and positive. On the other hand, the older adults in did not find the backchanneling version any more or less pleasant and acceptable. Still, backchanneling seemed to ease a perceived need for learning and support. It also elicited significantly more nonverbal backchanneling behaviours in the older adults themselves.

Comparing cross-generationally is limited because of the difference in research designs and data collection between Study 1 and Study 2. Even so, students who were more likely to be familiar with digital objects such as Zoom were aware of the presence of backchannelling and their evaluation was influenced by it. On the other hand, there was no marked difference in the evaluation of the older students who found the experience itself new. Unlike younger adults, though, older adults' behaviour was influenced: increased nonverbal backchanneling. The reason why is difficult to imagine, but it could relate to generational differences in understanding technology and experience with agents, i.e., older adults may unreflexively react to agentic technology as they would with any other (human, animal) agent. Notably, aside from technical difficulties for the older group, the online context did not appear to influence results, suggesting cross-generational viability if onboarding needs are met for older users.

*1) Limitations and Future Work*

We were not able to directly compare younger and older adults because we modified the research design, especially the type of data collection (e.g., backchannelling metrics). Future work will directly compare younger and older adults using the same procedures. We should also consider conducting a more tightly controlled study using external microphones, etc., so as to eliminate the influence of external factors as much as possible, or whether to make some allowances for practical reasons. These two perspectives are a trade-off, e.g., we can either let people use their own PCs as in Study 1 or provide PCs as in Study 2. We also need to improve the accuracy of the backchannelling. Currently, Bono-Bot only has "Minimal Response" backchannelling, such as "un-un." Given the results for older adults, it may also be effective to add non-verbal backchannelling to Bono-Bot or to augment "Listener Feedback" with longer sentences. Since we found that the robot elicited backchannelling, it may be necessary to examine the conditions under which the other types (Minimal response and Listener feedback) were elicited in with greater precision.

## VII. CONCLUSION

The study was conducted with both younger and older participants. However, older adults tended not to prefer to use the system. As the results show, difficulties and issues related to backchannelling and the online context were discovered. In addition, the accuracy and type of backchannelling was found to be more important than its presence alone. However, robotic backchannelling appeared to cause participants to also use backchannelling, especially non-verbal types. In the future, robots and conversations with robots, plus the use of the Coimagination Method platform, may be used by older adults, but further improvements are needed first. The effort is worth it: online systems for group interaction with intelligent agents have great potential to adapt and expand into the lives of people of all ages and could play a major role in supporting societies that are not adequately prepared for the rise in dementia and global pandemics such as COVID-19.

## APPENDIX: QUESTIONNAIRE

Q1. How about the experience of the conversations? [SUS items]
Q2. Do you think this experience was fun?
Q3. Do you think the facilitation robot listened well?
Q4. Do you think the facilitation robot fitted into the conversation well?
Q5. Do you want to be listened to by the facilitation robot?
Q6. Do you usually use this type of video call?
Q7. What were your impressions of this experience?
Q8. Do you have any recommendations to improve this experience?

## ACKNOWLEDGMENT

This research was partially supported by JSPS KAKENHI Grant Numbers JP18KT0035, JP19H01138, JP20H05022, JP20H05574, JP20K19471, JP22H04872, and JP22H00544, and the Japan Science and Technology Agency (JST grants JPMJCR20G1, JPMJPF2101, and JPMJMS2237). We also thank our participants and lab staff, especially Sachiko Iwata and Tomoko Suzuki, for recruitment and other support.

## REFERENCES

[1] M. Otake-Matsuura *et al.*, "Cognitive Intervention Through Photo-Integrated Conversation Moderated by Robots (PICMOR) Program: A Randomized Controlled Trial," *Front. Robot. AI*, vol. 8, p. 633076, Apr. 2021, doi: 10.3389/frobt.2021.633076.
[2] T. Suzuki and A. Miyata, "A Study of a Conversational Agent for Supporting Users' Understanding of Dementia Symptoms," 研究報告セキュリティ心理学とトラスト (SPT), vol. 2017-SPT-23, no. 9, pp. 1–5, May 2017.
[3] E. Marchetti, S. Grimme, E. Hornecker, A. Kollakidou, and P. Graf, "Pet-Robot or Appliance? Care Home Residents with Dementia Respond to a Zoomorphic Floor Washing Robot," in *CHI Conference on Human Factors in Computing Systems*, New Orleans LA USA: ACM, Apr. 2022, pp. 1–21. doi: 10.1145/3491102.3517463.
[4] M. Otake, "Development of Support Service for Prevention and Recovery from Dementia and Science of Lethe -Conversation Support Service for Social Interaction via Coimagination Method-," *Trans. Jpn. Soc. Artif. Intell. AI*, vol. Vol. 24, no. No. 6, pp. 569–576, 2006.
[5] K. Seaborn, T. Sekiguchi, S. Tokunaga, N. P. Miyake, and M. Otake-Matsuura, "Voice over body? Older adults' reactions to robot and voice assistant facilitators of group conversation," *Int. J. Soc. Robot.*, vol. 15, no. 2, pp. 143–163, Feb. 2023, doi: 10.1007/s12369-022-00925-7.
[6] 吉永敦征, "The Digital Devide of the Elderly and from the Elderly," 電子情報通信学会技術研究報告 IEICE Tech. Rep. 信学技報, vol. 118, no. 279, pp. 69–72, Nov. 2018.
[7] Y. Matsusaka, T. Tojo, and T. Kobayashi, "Conversation robot participating in group conversation," *IEICE Trans. Inf. Syst.*, vol. E86-D, no. 1, 2003.
[8] B. Mutlu, T. Shiwa, T. Kanda, H. Ishiguro, and N. Hagita, "Footing in human-robot conversations: how robots might shape participant roles using gaze cues," in *Proceedings of the 4th ACM/IEEE international conference on Human robot interaction - HRI '09*, La Jolla, California, USA: ACM Press, 2009, p. 61. doi: 10.1145/1514095.1514109.
[9] S. Fujie, Y. Matsuyama, H. Taniyama, and T. Kobayashi, "Conversation robot participating in and activating a group communication," in *Interspeech 2009*, ISCA, Sep. 2009, pp. 264–267. doi: 10.21437/Interspeech.2009-91.
[10] P. Althaus, H. Ishiguro, T. Kanda, T. Miyashita, and H. I. Christensen, "Navigation for human-robot interaction tasks," in *IEEE International Conference on Robotics and Automation, 2004. Proceedings. ICRA '04. 2004*, New Orleans, LA, USA: IEEE, 2004, pp. 1894-1900 Vol.2. doi: 10.1109/ROBOT.2004.1308100.
[11] M. Vázquez, E. J. Carter, B. McDorman, J. Forlizzi, A. Steinfeld, and S. E. Hudson, "Towards Robot Autonomy in Group Conversations: Understanding the Effects of Body Orientation and Gaze," in *Proceedings of the 2017 ACM/IEEE International Conference on Human-Robot Interaction*, Vienna Austria: ACM, Mar. 2017, pp. 42–52. doi: 10.1145/2909824.3020207.
[12] U. Takahisa, F. Tomo, S. Kurima, M. Takashi, and I. Hiroshi, "Promotion of Relationship Building between Users by Inducing Perspective-Taking: A Dialogue Strategy for Robots in Three Members' Dialogue," presented at the Human Interface Society, Aug. 2022. doi: 10.11184/his.24.3_167.
[13] H. Mori, "Dynamic aspects of aizuchi and its influence on the naturalness of dialogues," *Acoust. Sci. Technol.*, vol. 34, no. 2, pp. 147–149, 2013, doi: 10.1250/ast.34.147.
[14] 堀口順子, "コミュニケーションにおける聞き手の言語行動," 日本語教育 J. Jpn. Lang. Teach. 日本語教育学会学会誌委員会 編, no. 64, pp. 13–26, Mar. 1988.
[15] U. Hiroaki, T. Kazuaki, and N. Hideyuki, "A Research for the Automatic Nodding System based on Tuning Test," presented at the The 27th Annual Conference of the Japanese Society for Artificial Intelligence, 2013. [Online]. Available: https://doi.org/10.11517/pjsai.JSAI2013.0_1G53in
[16] K. Yuka, Y. Daisuke, T. Tsuyoshi, Y. Yuto, and D. Miwako, "Development of Dialog Interface for Elderly People -Voice Interface Robot using Dialogue Model between Nurses and Patients -," presented at the The 26th Annual Conference of the Japanese Society for Artificial Intelligence, 2012. [Online]. Available: https://doi.org/10.11517/pjsai.JSAI2012.0_2M24
[17] T. Yamaguchi, K. Inoue, Y. Koichiro, K. Takanashi, N. G. Ward, and T. Kawahara, "Generating a Variety of Backchannel Forms Based on Linguistic and Prosodic Features for Attentive Listening Agents," *Trans. Jpn. Soc. Artif. Intell.*, vol. 31, no. 4, p. C-G31_1-10, 2016, doi: 10.1527/tjsai.C-G31.
[18] S. Fujie, T. Kobayashi, and K. Fukushima, "A conversation robot with back-channel feedback function based on linguistic and nonlinguistic information," *Proc ICARA2004*, pp. 379–384, 2004.
[19] M. F. Jung, J. J. Lee, N. DePalma, S. O. Adalgeirsson, P. J. Hinds, and C. Breazeal, "Engaging robots: easing complex human-robot teamwork using backchanneling," in *Proceedings of the 2013 conference on Computer supported cooperative work - CSCW '13*, San Antonio, Texas, USA: ACM Press, 2013, p. 1555. doi: 10.1145/2441776.2441954.
[20] K. Kobayashi, K. Funakoshi, T. Komatsu, S. Yamada, and M. Nakano, "Improving User Experiences in Talking to Robots using ASE-based Back-channel Feedbacks," *Trans. Jpn. Soc. Artif. Intell.*, vol. 30, no. 4, pp. 604–612, 2015, doi: 10.1527/tjsai.30.604.
[21] B. John, "SUS: A 'Quick and Dirty' Usability Scale," in *Usability Evaluation In Industry*, 1996, p. 6.